\documentclass{article}

\PassOptionsToPackage{square,numbers}{natbib}
\usepackage[final]{neurips_2021}

\setcitestyle{square, numbers}

\usepackage[utf8]{inputenc} 
\usepackage[T1]{fontenc}    
\usepackage{hyperref}       
\usepackage{url}            
\usepackage{booktabs}       
\usepackage{amsfonts}       
\usepackage{nicefrac}       
\usepackage{microtype}      
\usepackage{xcolor}         
\usepackage[pdftex]{graphicx}

\title{SAR-based landslide classification pretraining leads to better segmentation}

\author{
  Vanessa B\"ohm\\
  University of California Berkeley \\
  United States \\
  \And
  Wei Ji Leong\\
  The Ohio State University \\
  United States \\
  \And
      Ragini Bal Mahesh\\
  German Aerospace Center DLR \\
  Germany \\
    \And
    Ioannis Prapas \\
  University of Valencia, Spain \\
  National Observatory of Athens, Greece \\
    \And
  Edoardo Nemni \\
  United Nations Satellite Centre (UNOSAT) \\
  Switzerland \\
  \And
  Freddie Kalaitzis \\
  University of Oxford \\
  United Kingdom \\
  \texttt{} \\
  \And
  Siddha Ganju \\
  NVIDIA \\
  United States \\
  \And
  Raul Ramos-Pollan \\
  Universidad de Antioquia \\
  Colombia \\
}

\begin{document}

\maketitle

\begin{abstract}
Rapid assessment after a natural disaster is key for prioritizing emergency resources. In the case of landslides, rapid assessment involves determining the extent of the area affected and measuring the size and location of individual landslides. Synthetic Aperture Radar (SAR) is an active remote sensing technique that is unaffected by weather conditions. Deep Learning algorithms can be applied to SAR data, but training them requires large labeled datasets. In the case of landslides, these datasets are laborious to produce for segmentation, and often they are not available for the specific region in which the event occurred. Here, we study how deep learning algorithms for landslide segmentation on SAR products can benefit from pretraining on a simpler task and from data from different regions. The method we explore consists of two training stages. First, we learn the task of identifying whether a SAR image contains any landslides or not. Then, we learn to segment in a sparsely labeled scenario where half of the data do not contain landslides. We test whether the inclusion of feature embeddings derived from stage-1 helps with landslide detection in stage-2. We find that it leads to minor improvements in the Area Under the Precision-Recall Curve, but also to a significantly lower false positive rate in areas without landslides and an improved estimate of the average number of landslide pixels in a chip. A more accurate pixel count allows to identify the most affected areas with higher confidence. This could be valuable in rapid response scenarios where prioritization of resources at a global scale is important.
We make our code publicly available at this \href{https://github.com/VMBoehm/SAR-landslide-detection-pretraining}{URL}.
\end{abstract}

\section{Introduction}
According to the United Nations Office for Disaster Risk Reduction, landslides have affected 4.8 million people and caused 18,414 deaths between 1998-2017\footnote{\url{https://www.preventionweb.net/files/61119_credeconomiclosses.pdf}}. Rising temperatures and climate change are projected to increase the likelihood of landslide triggered by atmospheric events~\cite{GarianoLandslidesChangingClimate2016}. Given these predictions, there is a growing need for timely and accurate landslide assessment methods that can inform decision makers and emergency responders.

In the aftermath of a natural disaster, optical satellite imagery is commonly used to map the extent of affected areas. Optical data are hindered by clouds and lack of daylight. This motivates the study of weather-independent Synthetic Aperture Radar (SAR) as a monitoring technique.

Deep Learning (DL) methods have been successfully applied for landslide detection (see~\cite{Tehrani_2022} for a review), including from SAR data~\cite{Nava_2022_1,Nava_2022_2}. Supervised deep learning algorithms require large labeled training datasets, which are costly to produce. 
Due to different terrain, vegetation coverage and geological characteristics, it is also unlikely that an algorithm trained on data from a certain area is readily applicable to another area. To date, very few landslide segmentation datasets exist, and they cover only small geographical areas (e.g.~\cite{Hokkaido_labels}). Larger landslide datasets contain only point labels (e.g.~\cite{Kaikoura_labels,bessette2019landslides}), where each landslide is marked by a point location instead of its contour, as these labels are faster and easier to annotate. 

Landslide detection from SAR data becomes more reliable with an increasing number of available satellite passes after an event~\cite{handwerger_generating_2022}. Combining data from several visits eases the distinction of random noise from the actual signal. A timely response will require inference from one satellite pass only. DL algorithms also struggle with highly unbalanced datasets, in which many of the images contain few or no landslides. Since the actual extent of an event is not known beforehand, such imbalances are difficult to avoid.

Here, we explore the use of pretraining to aid with the task of landslides mapping (segmentation) from SAR amplitude data when only one satellite pass is available. We propose a two-stage training set up. 
In the first stage, a network is trained on the task of identifying whether an image chip contains landslides or not. This task does not require segmentation labels which means that the network can be trained on almost any landslide dataset, and possibly with data from many different regions. 
After pretraining, the parameters of this network are kept fixed. We will refer to this network as the pretrained network.

In the second stage, we train a small network to predict landslide segmentation labels from a small labeled dataset. 
We test whether adding feature embedddings from the pretrained network is beneficial for this task.

\paragraph{Application context}
We explore pretraining as a means to enable landslide mapping from SAR imagery in realistic scenarios with scarce labels and a single satellite observation after the landslide event. The inferred maps can help with the allocation of resources in an emergency response.

\section{Data}
\paragraph{Landslide labels} We use two publicly available landslide datasets. The first dataset contains segmentation labels for landslides on the southern side of the island of Hokkaido, Japan, that were triggered by an Mw 6.6 earthquake on 6 September 2018~\cite{Hokkaido_labels}. The second dataset contains point labels, marking the centroids of landslides triggered by an Mw 7.8 earthquake on 14 November 2016 in Kaikoura, New Zealand~\cite{Kaikoura_labels}.

\paragraph{Input data} As input data we use PolSAR (polarimetric SAR) amplitudes from Sentinel-1 satellites that were acquired before and after the triggering events. The data was downloaded from the Microsoft Planetary Computing service\footnote{https://planetarycomputer.microsoft.com/}. For Hokkaido, we use data from two satellite passes on 2018/08/24, two weeks before the event, and on 2018/09/13, one week after the event. 
For Kaikoura we retrieve data from passes at 2016/08/23 (11 weeks before the event) and 2016/11/15 (the day after the event).
\paragraph{Chips generation} From labels and PolSAR data we generate data chips of size 128$\times$128 with 2 SAR polarimetry channels (VV + VH) for each of the Sentinel-1 revisits, and 1 channel for the labels. The pixel spatial resolution is 10m$\times$10m. In both datasets we randomly discard some of the chips that contains no landslides in order to obtain balanced datasets such that 50\% of the chips contain landslides. Then, we randomly split these datasets into a training set for pretraining training (50\%), a training set for the segmentation task (25\%), along with a validation and test set (12.5\% each). We ensure that each of these datasets are also balanced. The exact number of chips in each set is listed in the Appendix (Table~\ref{table:datasplits}). The Kaikoura dataset is larger (2174 chips) than the Hokkaido dataset (552 chips), but does not contain segmentation labels. Its point annotations (landslides centroids), however, can be used for the pretraining task. 

\section{Method}
\begin{figure}[htbp]
  \centering  \includegraphics[width=0.49\linewidth]{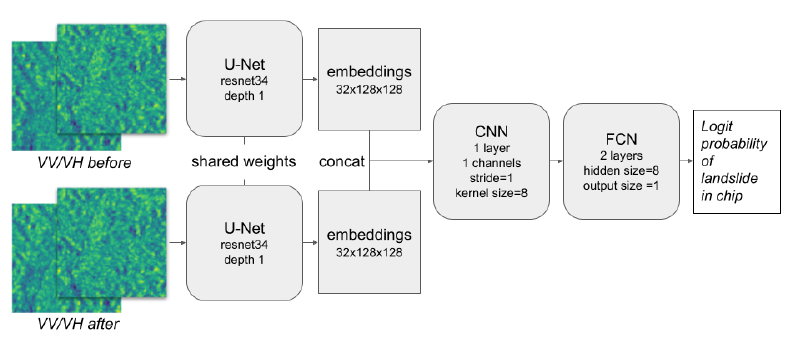}
  \includegraphics[width=0.49\linewidth]{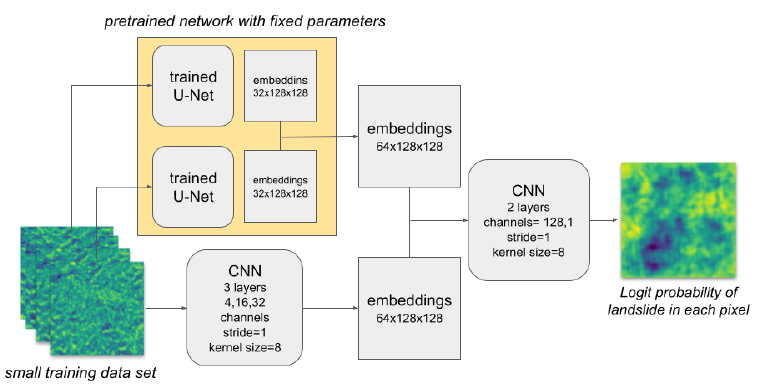}
  \caption{Stage-1 and stage-2 training pipelines (left and right). 
  }
  \label{fig:networks}
  \vspace{-0.3cm}
\end{figure}

We aim to perform landslide segmentation from PolSAR data with sparsely labeled image chips and a single satellite pass after the landslide event. Our method is inspired by transfer learning techniques, in which a network trained on a large dataset with abundant labels, is reused to aid with a different but related task (see~\cite{Weiss2016ASO} for a review). Figure~\ref{fig:networks} shows the architecture for the two stages of our method.
Our pretraining task (stage-1) is a classification task which requires hand-annotated point labels. These labels, however, are much faster to generate than segmentation labels. In addition, any type of label (point or segmentation) can be used for this pretraining task, increasing the amount of available training data significantly.

In the first stage, we train a network to predict whether a given chip contains any landslides or not. Our setup is inspired by Siamese networks~\cite{BromleyBBGLMSS93,Chicco2021}: The network is presented with PolSAR images before and after the event separately. Then, the output embeddings of size 64x128x128 are concatenated and fed to a two-layer fully connected network head to compute a scalar output. The final output is interpreted as the logit probability of the data chip containing a landslide.

We use a U-Net~\cite{RonnebergerUNetConvolutionalNetworks2015} architecture with a ResNet-34 encoder for this task (21M parameters). We find that a simple CNN yields slightly lower accuracies. We note that this learning task is remarkably challenging, due to the noise in the data. Simple network architectures regress to the mean, while larger networks quickly overfit. The best models only reach a validation accuracy of 0.6. Performing the same task with inputs averaged over 5 revisits after the event increases the validation accuracy 0.8.
After training, we discard the network head and freeze the parameters in the pretrained U-Net.

In stage-2 the input data chip is fed both through a 3-layer CNN to produce embeddings, as well as through the pretrained Stage-1 network. The embeddings of Stage-1 and the CNN are then combined and further processed with a two-layer CNN. The output is interpreted as the logit probability of each pixel belonging to a landslide. This network now has around 1M parameters.
  

\subsection{Training procedures}
The U-Net in stage-1 is trained on a standard cross entropy loss with a learning rate of 0.001, a batch size of 32 and an Adam optimizer with default parameters\footnote{\texttt{https://pytorch.org/docs/stable/generated/torch.optim.Adam.html}}. We stop the training with an early stopping criterion of no improvement in validation loss for 50 epochs. During training we save model checkpoints for the training steps with lowest validation accuracy. We pretrain two models---one on the Hokkaido and one on the Kaikoura datasets---and we compare their performance. We note that the training chips used for pretraining on the Hokkaido data are not used in the segmentation task.

The segmentation network in stage-2 is trained on the Hokkaido data segmentation split. We use the same batchsize, optimizer and learning rate as for stage-1, but train on the dice loss~\cite{diceloss}, since the dataset is highly imbalanced. 

We perform ablation studies with different training set sizes, ranging from 2 to 110 chips. As a baseline we also train models of the same architecture but without adding the embeddings from the pretrained model.  
Each experiment is repeated three times with three different random seeds for 5, 10, 20 and 110 training chips, and 5 times with five different random seeds for 2 training chips. If less than the full training set (110 chips) is used, we randomly sub-sample the training set, meaning that each run uses a different set of training chips. During training we save the 5 model checkpoints with highest Area under Recall-Precision Curve on the validation dataset in each training. With this procedure we end up with a total of 3x5 model checkpoints for each training set size ((5x5) for 2 chips). Metrics are computed from averages over these checkpoints.

\section{Results}
We measure performance in terms of the following metrics: 1) The Area under the Precision-Recall Curve (APRC), which is a robust measure for imbalanced datasets and accounts for all thresholds to distinguish between positive and negative predictions. 2) Counting errors (defined below) that measure the difference in the number of predicted landslide pixels and number of true landslide pixels in each chip. This metric is insensitive to the overlap between predicted and true landslide pixels. Instead, it provides a measure of how well the network estimates the landslide density in each chip. All reported metrics are measured on the test set.

\begin{figure}
  \centering
  \includegraphics[width=0.4\linewidth]{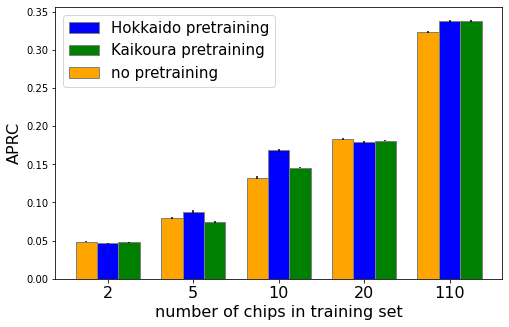}
  \caption{Area under Precision-Recall curve (APRC $\uparrow$) measured on the test dataset after training on 2, 5, 10,20 and 110 training chips. Results are averaged over training runs on different random seeds and different model checkpoints saved during training. Pretraining on either datasets does not yield any significant improvements in APRC.}
  \label{fig:APRC}
\vspace{-0.5cm}
\end{figure}

To increase the robustness of the results, we compute metrics on average model predictions, where the average is taken over outputs from models trained with different random seeds and different model checkpoints saved during training. The standard deviation of the mean is reported as error bars. 

Results for APRC are shown in Figure~\ref{fig:APRC}. 
For reference, the APRC for a random pixel assignment in this dataset is 0.032. We find that pretraining only results in negligible improvements in APRC over no pretraining in all scenarios. Pretraining on the Hokkaido dataset yields similar APRCs as pretraining on the Kaikoura dataset.

To gain further insight, we also compute the error in the total number of landslide pixels in a chip (using a decision threshold of p=0.5). We define two error metrics, 
\begin{equation}
\label{eq:L1}
    \Delta \mathrm{L}^j_1 = \left|  \sum_i \tilde{y}^j_i-\sum_i y^j_i \right|\,\,\,\,\,\,\,\,\mathrm{and}\,\,\, \quad \Delta \mathrm{count}^j = \sum_i \tilde{y}^j_i-\sum_i y^j_i,
\end{equation}
where $j$ indexes chips in the test set and $i$ indexes pixels in each chip. We use $y$ to denote the true labels and $\tilde{y}$ to denote predicted labels.
For chips containing no landslides $\Delta \mathrm{L}_1$ measures the average number of false positives. $\Delta\mathrm{count}$ measures if a model over- or underestimate the counts.

\begin{figure}[htbp]
  \centering
  \includegraphics[width=0.32\linewidth]{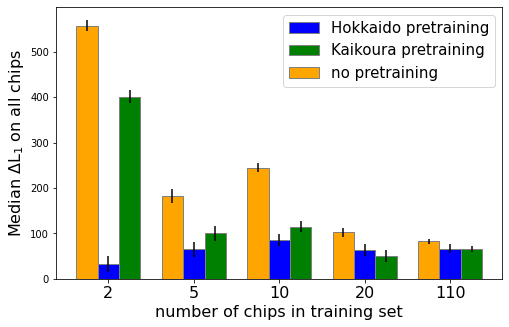}
  \includegraphics[width=0.32\linewidth]{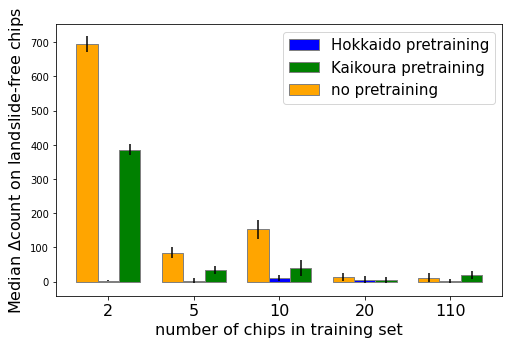}
  \includegraphics[width=0.32\linewidth]{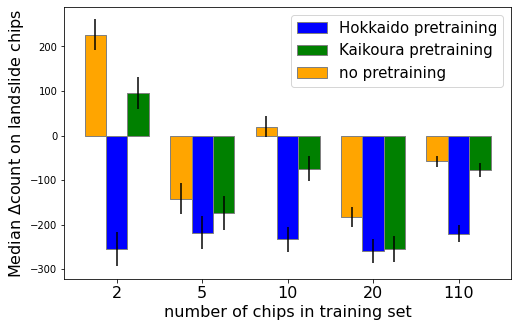}
  \caption{Comparison of landslide pixel miscounts on the test set in models that use embeddings from the pretrained U-Net and baseline models without pretraining. From left to right, we show absolute differences in counts ($\Delta \mathrm{L}_1$) in all test chips, differences in counts ($\Delta\mathrm{count}$) on chips that contain no landslides and differences in counts in chips that contain landslides.}
  \label{fig:l1_scores}
\vspace{-0.2cm}
\end{figure}

We report the median of miscounts (Eq.~\ref{eq:L1}) over all chips in the test set in the left panel of Figure~\ref{fig:l1_scores}. We use the median since it is less sensitive to outliers than the mean. Using mean values, however, does not alter the conclusions. Adding embeddings from the pretrained networks results in smaller counting errors in all cases. By looking at the $\Delta\mathrm{count}$ metric, we see that this is driven by a smaller number of false positives in chips with no landslides (middle panel). On chips containing landslides (right panel) the miscount is generally larger when embeddings from the pretrained network are used. The negative sign of the count difference indicates that these models are prone to underestimate the number of landslide pixels.  

All findings are consistent between pretraining on the Hokkaido dataset and pretraining on the Kaikoura dataset, but more pronounced for pretraining on Hokkaido. This suggests that the pretraining has indeed added information. We show representative visual examples of these findings in the appendix.

\section{Conclusion}
We have explored the usefulness of pretraining for the realistic and extremely challenging task of landslide segmentation from SAR polarimetry data if only few satellite passes are available and the training data is sparse. 

We find that pretraining on the task of identifying whether a data chip contains landslides only results in a very small, but hardly significant improvement in the APRC metric on the segmentation task. However, it improves estimates of the average number of landslide pixels in a chip. A more accurate pixel count allows to identify the most affected areas with higher confidence. This can be beneficial in emergency response situations, where prioritization of resources is crucial. The benefit is present even if pretraining was performed on data from a different area with different characteristics.

In the future, calibration techniques (e.g. calibrating the threshold probability for identifying pixels as landslides) and additional input data, such as elevation data, could be used to refine and improve this ansatz.

\begin{ack}
This work has been enabled by the Frontier Development Lab Program (FDL).
FDL is a collaboration between SETI Institute and Trillium Technologies Inc., in partnership with Department of Energy (DOE) , National Aeronautics and Space Administration (NASA), and U.S. Geological Survey (USGS).
The material is based upon work supported by NASA under award No(s) NNX14AT27A.
Any opinions, findings, and conclusions or recommendations expressed in this material are those of the authors and do not necessarily reflect the views of the National Aeronautics and Space Administration.

We thank Aaron Pina, our contact from NASA who helped shape this research.
He also facilitated communication with experts from NASA and ASF, namely Gerald Bawden (NASA), Eric Fielding (NASA), Alex Handwerger (NASA), Erika Podest (NASA), Franz Meyer (ASF), Forrest Williams (ASF).
We also thank our FDL reviewers Brad Neuberg (Planet), Ronny Hänsch (DLR), Erika Podest (NASA), Ioannis Papoutsis (National Observatory of Athens) who provided great feedback on specificities of ML for SAR.

\end{ack}

\bibliography{references.bib}

\begin{thebibliography}{13}
\providecommand{\natexlab}[1]{#1}
\providecommand{\url}[1]{\texttt{#1}}
\expandafter\ifx\csname urlstyle\endcsname\relax
  \providecommand{\doi}[1]{doi: #1}\else
  \providecommand{\doi}{doi: \begingroup \urlstyle{rm}\Url}\fi

\bibitem[Gariano and Guzzetti(2016)]{GarianoLandslidesChangingClimate2016}
Stefano~Luigi Gariano and Fausto Guzzetti.
\newblock Landslides in a changing climate.
\newblock \emph{Earth-Science Reviews}, 162:\penalty0 227--252, November 2016.
\newblock ISSN 00128252.
\newblock \doi{10.1016/j.earscirev.2016.08.011}.
\newblock URL
  \url{https://linkinghub.elsevier.com/retrieve/pii/S0012825216302458}.

\bibitem[Tehrani et~al.(2022)Tehrani, Calvello, Liu, Zhang, and
  Lacasse]{Tehrani_2022}
Faraz~S. Tehrani, Michele Calvello, Zhongqiang Liu, Limin Zhang, and Suzanne
  Lacasse.
\newblock Machine learning and landslide studies: recent advances and
  applications.
\newblock \emph{Nat Hazards}, jun 2022.
\newblock \doi{10.1007/s11069-022-05423-7}.
\newblock URL \url{https://doi.org/10.1007%2Fs11069-022-05423-7}.

\bibitem[Nava et~al.(2022)Nava, Bhuyan, Meena, Monserrat, and
  Catani]{Nava_2022_1}
Lorenzo Nava, Kushanav Bhuyan, Sansar~Raj Meena, Oriol Monserrat, and Filippo
  Catani.
\newblock Rapid mapping of landslides on {SAR} data by attention u-net.
\newblock \emph{Remote Sensing}, 14\penalty0 (6):\penalty0 1449, mar 2022.
\newblock \doi{10.3390/rs14061449}.
\newblock URL \url{https://doi.org/10.3390%2Frs14061449}.

\bibitem[{Nava} et~al.(2022){Nava}, {Monserrat}, and {Catani}]{Nava_2022_2}
Lorenzo {Nava}, Oriol {Monserrat}, and Filippo {Catani}.
\newblock {Improving Landslide Detection on SAR Data Through Deep Learning}.
\newblock \emph{IEEE Geoscience and Remote Sensing Letters}, 19:\penalty0
  3127073, January 2022.
\newblock \doi{10.1109/LGRS.2021.3127073}.

\bibitem[Zhang et~al.(2019)Zhang, Li, Wang, and Iio]{Hokkaido_labels}
Shuai Zhang, Ran Li, Fawu Wang, and Akinori Iio.
\newblock Characteristics of landslides triggered by the 2018 hokkaido eastern
  iburi earthquake, north japan.
\newblock \emph{Zenodo}, Feb 2019.
\newblock \doi{10.5281/zenodo.2577300}.

\bibitem[Massey et~al.(2021)Massey, Townsend, Rosser, Morgenstern, Jones,
  Lukovic, and Davidson]{Kaikoura_labels}
C.~Massey, D.~Townsend, B.~Rosser, R.~Morgenstern, K.~Jones, B.~Lukovic, and
  J.~Davidson.
\newblock Version 2.0 of the landslide inventory for the mw 7.8 14 november
  2016, kaikōura earthquake, 2021.

\bibitem[Bessette-Kirton et~al.(2019)Bessette-Kirton, Cerovski-Darriau, Schulz,
  Coe, Kean, Godt, Thomas, and Hughes]{bessette2019landslides}
Erin~K Bessette-Kirton, Corina Cerovski-Darriau, William~H Schulz, Jeffrey~A
  Coe, Jason~W Kean, Jonathan~W Godt, Matthew~A Thomas, and K~Stephen Hughes.
\newblock Landslides triggered by hurricane maria: Assessment of an extreme
  event in puerto rico.
\newblock \emph{GSA Today}, 29\penalty0 (6):\penalty0 4--10, 2019.

\bibitem[Handwerger et~al.(2022)Handwerger, Huang, Jones, Amatya, Kerner, and
  Kirschbaum]{handwerger_generating_2022}
Alexander~L. Handwerger, Mong-Han Huang, Shannan~Y. Jones, Pukar Amatya,
  Hannah~R. Kerner, and Dalia~B. Kirschbaum.
\newblock Generating landslide density heatmaps for rapid detection using
  open-access satellite radar data in {Google} {Earth} {Engine}.
\newblock \emph{Natural Hazards and Earth System Sciences}, 22\penalty0
  (3):\penalty0 753--773, March 2022.
\newblock ISSN 1561-8633.
\newblock \doi{10.5194/nhess-22-753-2022}.
\newblock URL \url{https://nhess.copernicus.org/articles/22/753/2022/}.
\newblock Publisher: Copernicus GmbH.

\bibitem[Weiss et~al.(2016)Weiss, Khoshgoftaar, and Wang]{Weiss2016ASO}
Karl~R. Weiss, Taghi~M. Khoshgoftaar, and Dingding Wang.
\newblock A survey of transfer learning.
\newblock \emph{Journal of Big Data}, 3:\penalty0 1--40, 2016.

\bibitem[Bromley et~al.(1993)Bromley, Bentz, Bottou, Guyon, LeCun, Moore,
  S{\"{a}}ckinger, and Shah]{BromleyBBGLMSS93}
Jane Bromley, James~W. Bentz, L{\'{e}}on Bottou, Isabelle Guyon, Yann LeCun,
  Cliff Moore, Eduard S{\"{a}}ckinger, and Roopak Shah.
\newblock Signature verification using {A} "siamese" time delay neural network.
\newblock \emph{Int. J. Pattern Recognit. Artif. Intell.}, 7\penalty0
  (4):\penalty0 669--688, 1993.
\newblock \doi{10.1142/S0218001493000339}.
\newblock URL \url{https://doi.org/10.1142/S0218001493000339}.

\bibitem[Chicco(2021)]{Chicco2021}
Davide Chicco.
\newblock \emph{Siamese Neural Networks: An Overview}, pages 73--94.
\newblock Springer US, New York, NY, 2021.
\newblock ISBN 978-1-0716-0826-5.
\newblock \doi{10.1007/978-1-0716-0826-5_3}.
\newblock URL \url{https://doi.org/10.1007/978-1-0716-0826-5_3}.

\bibitem[Ronneberger et~al.(2015)Ronneberger, Fischer, and
  Brox]{RonnebergerUNetConvolutionalNetworks2015}
Olaf Ronneberger, Philipp Fischer, and Thomas Brox.
\newblock U-{Net}: {Convolutional} {Networks} for {Biomedical} {Image}
  {Segmentation}, May 2015.
\newblock URL \url{http://arxiv.org/abs/1505.04597}.
\newblock arXiv:1505.04597 [cs].

\bibitem[Milletari et~al.(2016)Milletari, Navab, and Ahmadi]{diceloss}
Fausto Milletari, Nassir Navab, and Seyed{-}Ahmad Ahmadi.
\newblock V-net: Fully convolutional neural networks for volumetric medical
  image segmentation.
\newblock In \emph{Fourth International Conference on 3D Vision, 3DV 2016,
  Stanford, CA, USA, October 25-28, 2016}, pages 565--571. {IEEE} Computer
  Society, 2016.
\newblock \doi{10.1109/3DV.2016.79}.
\newblock URL \url{https://doi.org/10.1109/3DV.2016.79}.

\end{thebibliography}
\clearpage
\appendix
\section{Additional figures and tables} 
\newcommand{\hbAppendixPrefix}{A}
\renewcommand{\thefigure}{\hbAppendixPrefix\arabic{figure}}
\setcounter{figure}{0}

\renewcommand{\thetable}{\hbAppendixPrefix\arabic{table}}
\setcounter{table}{0}
\begin{figure}[htp]
  \centering
  \begin{minipage}{0.9\linewidth}
      
  \includegraphics[width=0.49\linewidth]{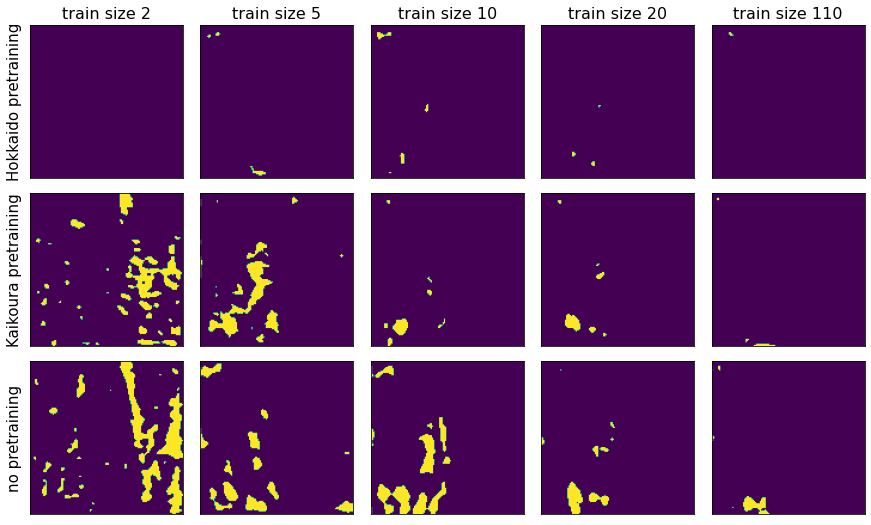}
  \includegraphics[width=0.49\linewidth]{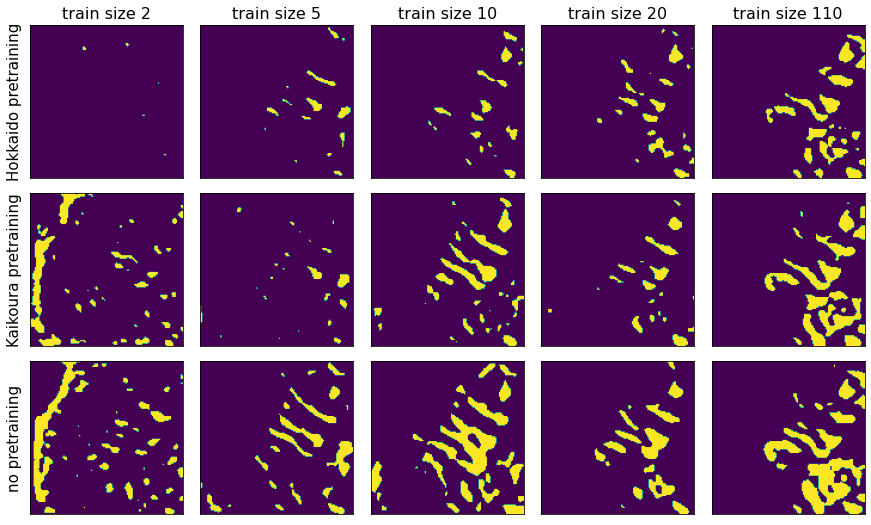}
  \end{minipage}
  \begin{minipage}{0.09\linewidth}
  \includegraphics[width=\linewidth]{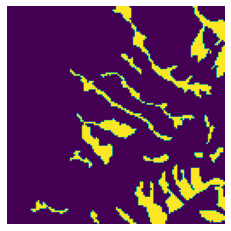}
  \end{minipage}
  \caption{Model predictions for segmentation maps (threshold=0.5) on a test data chip containing no landslides (left) and a data chip containing landslides (right). The true labels for the landslide chip are shown on the far right. The predictions are inferred from averaging over predictions from models trained with different random seeds as described in the text. Models with access to pretrained embeddings have less false positives in chips without landlides but also tend to underestimate the landslide density, if landslides are present. 
  }
  \label{fig:underestimation}
\end{figure}  

\begin{table}[h]
  \caption{Number of chips reserved for each task in each dataset. The Kaikoura dataset is much larger, but contains point labels, not segmentation labels. }
  \label{table:datasplits}
  \centering
  \begin{tabular}{l|lllll}
    \toprule
    dataset &  pretraining (stage-1) & segment. training (stage-2) & validation & testing & total \\
    \midrule
 Hokkaido & 331	& 110  & 55	& 56 & 552\\
 Kaikoura & 1630 & - & 272	& 272 & 2174\\
    \bottomrule
  \end{tabular}
\end{table}

\end{document}